\documentclass[conference]{IEEEtran}
\usepackage[utf8]{inputenc}
\usepackage{graphicx}
\usepackage[flushleft]{threeparttable}
\usepackage[ruled,vlined,linesnumbered]{algorithm2e}

\begin{document}

\title{Adaptive Precision Training for Resource Constrained Devices}

\author{
\IEEEauthorblockN{Tian HUANG\IEEEauthorrefmark{1}, 
Tao LUO\IEEEauthorrefmark{2}, 
and 
Joey Tianyi ZHOU\IEEEauthorrefmark{3}}
\IEEEauthorblockA{Agency for Science, Technology and Research, Singapore\\
\IEEEauthorrefmark{1}huang\_tian@ihpc.a-star.edu.sg,
\IEEEauthorrefmark{2}luo\_tao@ihpc.a-star.edu.sg,
\IEEEauthorrefmark{3}Joey\_Zhou@ihpc.a-star.edu.sg
}
}

\maketitle

\begin{abstract}
Learn in-situ is a growing trend for Edge AI. Training deep neural network (DNN) on edge devices is challenging because both energy and memory are constrained. Low precision training helps to reduce the energy cost of a single training iteration, but that does not necessarily translate to energy savings for the whole training process, because low precision could slows down the convergence rate. One evidence is that most works for low precision training keep an fp32 copy of the model during training, which in turn imposes memory requirements on edge devices. In this work we propose Adaptive Precision Training. It is able to save both total training energy cost and memory usage at the same time. We use model of the same precision for both forward and backward pass in order to reduce memory usage for training. Through evaluating the progress of training, APT allocates layer-wise precision dynamically so that the model learns quicker for longer time. APT provides an application specific hyper-parameter for users to play trade-off between training energy cost, memory usage and accuracy. Experiment shows that APT achieves more than 50\% saving on training energy and memory usage with limited accuracy loss. 20\% more savings of training energy and memory usage can be achieved in return for a 1\% sacrifice in accuracy loss.
\end{abstract}

\section{Introduction}

Deep learning-based Edge AI, as a growing trend, have started to change many aspects of people’s lives. Well trained deep learning models are able to achieve record-breaking predictive performance \cite{krizhevsky2012imagenet, hinton2012deep}, but we have witnessed increasing demand for the model to learn in-situ, for the purpose of personalisation or adaptation to evolving environment. Comparing to cloud-based deep learning framework, training a deep learning model on edge device reflects the very idea of edge computing, and benefits from energy savings in data transmission, low response latency and enhanced privacy.

Train deep learning model on edge device is challenging because training DNNs usually involves energy-intensive devices such as GPU and CPU with high precision (float32) processing units and abundant memory. For example, neural architecture search on ImageNet \cite{cai2017deep} would consume 10,000 GPU hours and tons of memory. An edge device such as a smart phone could easily draining out its battery before finishing a few training epochs. Training is memory demanding than inference, as the precision demand for back propagation is usually higher than inference \cite{hubara2017quantized} \cite{zhou2016dorefa}. Many works \cite{hubara2017quantized, li2016ternary} only focus on compressing network for inference, while still keeping high precision during training. The gap between limited resource on edge devices and the growing demand on training is widening, as researchers strive for better accuracy and propose larger models for solving more complicated tasks \cite{kaiser2017one}.

Low precision training reduces the energy cost of a single training iteration, but that does not necessarily translate to energy savings for the whole training process, because low precision could slows down the convergence rate. \cite{hubara2017quantized} A prolonged training process could leads to a worse battery situation. Many works keep an fp32 copy of their model during training \cite{hubara2017quantized, li2016ternary, zhu2016trained, zhou2016dorefa, wen2017terngrad, wang2019e2}, which in turn imposes further memory requirements on edge devices. 

In this paper we consider the training of a DNN on energy and memory constrained devices. Our goal is to \textbf{reduce the total energy and memory cost for the training}, We are inspired by \cite{gupta2015deep} and find that one can train a model with low precision at the beginning of training process, enjoying the energy and memory savings from low precision. As the training curve comes to its plateau, adding more precision help the model approach higher accuracy with fewer training epochs and less energy.

Based on this inspiration, we propose a heuristic method called \textbf{Adaptive Precision Training (APT)}. In APT, We use per-layer metrics that indicate how effective a layer learns with given precision. By evaluating the metrics, APT adjusts layer-wise precision dynamically throughout the training process. Below shows a few features that sets APT unique among other training methods:

\begin{itemize}
  \item APT adjusts layer-wise precision dynamically (i.e. using different bitwidth for different layer in different period of training).
  \item In training, we use a model of the same precision for both forward and backward propagation in order to save memory usage for training.
  \item APT provides an application specific hyper-parameter that achieves trade-off between training energy, memory usage and accuracy.
\end{itemize}
 
APT finds the layer-wise precision configuration that help the model learn effectively on the fly, It benefits edge devices that have to learn in-situ frequently after deployment. Experiments show that with APT, a model starts in low precision is able to learn as fast as, sometimes even faster than, its fp32 counterpart. APT achieves more than 50\% saving on training energy and memory usage with limited accuracy loss. By tuning the application specific hyper-parameter, one can achieve 20\% more savings of training energy and memory usage in return for a 1\% sacrifice in accuracy loss.

\section{Related Work}

We mainly focus on adaptive layer-wise precision for training. Orthogonal and complementary techniques for reducing complexity like network compression, pruning \cite{han2015deep, zhou2017incremental} and compact architectures \cite{howard2017efficient} are impressively efficient but outside the scope this paper.

\textbf{Low Precision for Forward Propagation} Many works explores the capability of fixed-point representation for forward prorogation (FPROP) \cite{jacob2018quantization, wang2019haq, wu2018mixed, yang2019quantization}. These methods keep a floating-point copy of model parameters in the backward propagation (BPROP), which places additional demand on memory usage and data movement during training. These floating-point model parameters are quantised into fixed-point numbers during FPROP.

\textbf{Quantisation Aware Training} Training with quantised gradients has been well studied in the distributed learning, whose main motivation is to reduce the communication cost during gradient aggregations between workers \cite{bernstein2018signsgd, wen2017terngrad}. Our goal setting are different because an edge device may not have access to remote support and has to learn in-situ by itself. DoReFa-Net \cite{zhou2016dorefa} quantise gradients to low-bitwidth fixed-point numbers in the BPROP. TernGrad \cite{wen2017terngrad} quantises gradients to ternary values in the BPROP. But the weights are stored and updated in floating-point representations for these works. WAGE \cite{wu2018training} provided a lossy information compression technique and manage to operates BPROP in 8-bit. The goal setting of this work is different from us, as it focuses on reducing the energy cost for single training iteration.

\textbf{Adaptive Training} There are a few works that quantise weight and activation with layer-wise precision. \cite{wang2019haq, wu2018mixed, yang2019quantization} explore the space of layer-wise bitwidth for energy efficient inference. \cite{zhang2020precision} learns quantisation parameters through finding gradient of these parameters. Adding more learnable parameters could introduce more uncertainty of the convergence of training curve. It also introduce more energy cost for learning more parameters. Zhang et al. \cite{zhang2019adaptive} also emphasise on energy saving for single training iterations and store and update weights in fp32 format in BPROP. 

\section{Adaptive Precision Training}

In this work, we apply a widely used quantisation scheme \cite{jacob2018quantization} to convert floating point numbers to into integers. The scheme is equivalent to an \emph{affine mapping} of integers $q$ to real numbers $r$. The formalised equation is shown as follows $r=S(q-Z)$. $S$ and $Z$ represent the scale and zero point of a group of values, or a tensor. All values in a tensor share one $S$ and $Z$. Different tensors have their own $S$ and $Z$. For $k$-bit quantisation, $q$ has $2^k$ possible discrete states. 

\subsection{Quantisation Underflow}

Low precision integer arithmetic is cheaper than a floating point arithmetic in terms of energy and memory, but quantisation magnifies the underflow issue. \cite{krishnamoorthi2018quantizing} Equation \ref{eq:weight_update} shows the update process of a weight in fp32 format:

\begin{equation}
\label{eq:weight_update}
w_{ij} := w_{ij} - lr * g_{ij}
\end{equation}

$w_{ij}$ is the $j$-th weight of $i$-th layer, $lr$ is the learning rate, $g_{ij}$ is the corresponding gradient of the $w_{ij}$. Ideally, at each training step, a weight changes by $lr * g_{ij}$. As we apply $k$-bit quantisation to a tensor of weights, too small changes cannot be represented by the weight. We refer to this minimum resolution as $\epsilon$, which is formalised as follow:

\begin{equation}
\label{eq:epsilon}
\epsilon _i= \frac{max(W_i)-min(W_i)}{2^k-1}
\end{equation}

$W_i$ is a tensor of weights of $i$-th layer. $k$ is the precision, or the bitwidth for the tensor. The updates of a weight $w_{ij}$ is regulated by $\epsilon _i$, which can be formalised as the following equation:

\begin{equation}
w_{ij} := w_{ij} - \left \lfloor \frac{lr * g_{ij}}{\epsilon _i} \right \rfloor * \epsilon _i
\label{eq:weight_update_condition}
\end{equation}

As a result, $lr * g_{ij}$ is quantised to discrete states, which is equivalent to that of $w_{ij}$. $lr * g_{ij}$ has to be at least bigger than $\epsilon _i$, otherwise quantisation underflow happens. For high precision training (e.g. 32-bit) $\epsilon$ is small, $lr * g_{ij}$ is bigger than $\epsilon _i$ for most of cases. As the precision decreases, $\epsilon _i$ increases, quantisation underflow happens more often, putting higher resistance on weights updating. 

Gradient is the horse power that drives the training process forward. Low-precision training put restrictions on the horse power and slow down the training process. We have the observation that in low precision network, gradients vanishes quicker than that of a network with higher precision. This is because quantisation underflow happens more often in low precision model and guides the model into local minima. Some of layers have larger $\epsilon$ and are harder to be updated than others. As the training loss decreases, These layers with larger $\epsilon$ suffer more quantisation underflow than before, stepping onto a slippery slope. As the training process moves forward, quantisation underflow freezes more parameters, driving the training into a dead state.

\subsection{Metrics for Underflow}

Many works \cite{hubara2017quantized, li2016ternary, zhu2016trained, zhou2016dorefa, wen2017terngrad, wang2019e2, zhang2019adaptive} keep an fp32 copy of all parameters in order to prevent the above-mention situation from happening, in the cost of memory usage, energy on data movement and additional training iterations. We believe there are low precision configurations that is able to save training energy and memory usage at the same. The key is to understand how easy quantisation underflow happens to a layer, I propose a metric called $Gavg$, which is formalised as Equation \ref{eq:gavg_metric}.

\begin{equation}
\label{eq:gavg_metric}
Gavg_i = \frac{1}{N_i}\sum_{j=0}^{N_i} \left | \frac{g_{ij}}{\epsilon _i} \right |
\end{equation}

$N_i$ represents the number of parameters in a tensor, $Gavg_i$ indicates how large a gradient is related to the minimum resolution of parameters in $i$-th layer. The larger the $Gavg$ is, the less likely the parameters remains unchanged during back propagation. If $Gavg$ approaches zero, that means the layer suffers serious quantisation underflow problem and does not update for most of times. The precision is the key to prevent $Gavg$ from approaching zero. A higher precision leads to lower $\epsilon$ and higher $Gavg$, which means the parameters are easier to update.  

Although we use weight to describe the concept of quantisation underflow and metrics, $Gavg$ applies to other parameters that need to be learned during training, e.g. bias, the clipping point of activation and gradient. In the metric $Gavg$, we do not include other factors like learning rate or momentum in the metrics so that user can still use training tricks or sophisticated optimisers over our training method.

\subsection{Precision Adjustment Policy}

The metric $Gavg$ is a good measure for balancing requirements between training energy, memory usage, and accuracy. It is well known that precision is directly related to the energy, memory and accuracy. Using fixed precision across the whole network may not meets the requirement of each layer. On comparison, a single threshold for $Gavg$ is able to produce different precision configuration according to the distribution of parameters and their gradients.

We propose a precision adjustment policy based on the metric $Gavg$. We start with a naive policy, which is to make sure that $Gavg$ of all layers are within a pre-defined range. A C-styled description of the policy is presented below.

\begin{algorithm}
\SetAlgoNoLine
\SetKwFor{For}{for (}{) $\lbrace$}{$\rbrace$}
\SetKwFor{If}{if (}{) $\lbrace$}{$\rbrace$}
\SetKwInOut{Input}{input}
\SetKwInOut{Output}{output}
\Input{$k_{0\ldots M-1}$, $Gavg_{0\ldots M-1}$, $T_{min}$, $T_{max}$, }
\Output{$k_{0\ldots M-1}$}
 \For {$i = 0;\ i < M;\ i = i + 1$}{
  \If{$Gavg_i < T_{min}$ \&\& $k_i < 32$}{
   $k_i := k_i + 1$\;
   }
   \If{$Gavg_i > T_{max}$ \&\& $k_i > 2$}{
   $k_i := k_i - 1$\;
   }
 }
 \Return{$k_{0\ldots M-1}$}\;
 \caption{Precision Adjustment Policy}
 \label{alg:policy}
\end{algorithm}

In Algorithm \ref{alg:policy},  $k_{0\ldots M-1}$ and $Gavg_{0\ldots M-1}$ represents the precision and metrics of all $M$ layers. $T_{min}$, $T_{max}$ are the upper and lower limit. Algorithm \ref{alg:policy} increase the precision of a layer when its $Gavg < T_{min}$ and decrease the precision when $Gavg > T_{max}$. The lower limit ensures all layers learn effective, whereas the upper limit is for saving energy cost and memory usage on those parameters that are very easy to update (e.g. due to its small range or large gradients).

\begin{algorithm}
\SetAlgoNoLine
\SetKwFor{For}{for (}{) $\lbrace$}{$\rbrace$}
 Initialise all layers of a model with low precision, e.g. $k=6$;\\
 \For {$epoch = 0;\ epoch < 200;\ epoch++$}{
    \For {$iter = 0;\ iter < len(train\_loader);\ iter++$}{
        Forward propagation;\\
        Backward propagation;\\
        \If{$iter \% INTERVAL == 0$}{
            Evaluate $Gavg$ using Equation \ref{eq:gavg_metric};\\
            Moving average on $Gavg$;\\
        }
    }
    Adjust model precision using Algorithm \ref{alg:policy};\\
 }
 \caption{Training with APT}
 \label{alg:training}
\end{algorithm}

A typical workflow of APT is described in Algorithm \ref{alg:training}. A training starts with a low-precision model. The metrics evaluation happens inside each training epoch. The precision adjustment happens between training epochs. $Gavg$ does not have to be calculated for each training iteration. A few times in each epoch suffice to give a profile of $Gavg$ for precision adjustment. 

\begin{figure}
\centering
\includegraphics[width=0.9\linewidth]{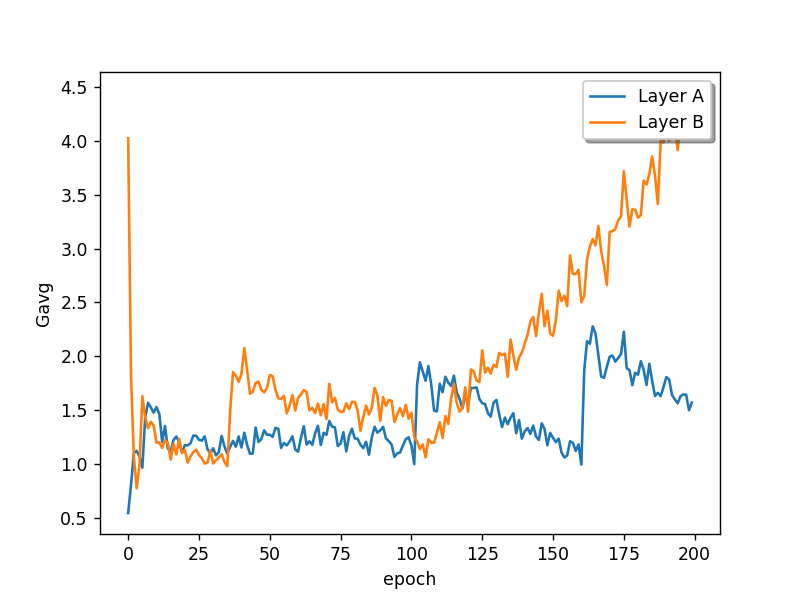}
\caption{Gavg v.s. Epoch for two layer}
\label{fig:gavgvsepoch}
\end{figure}

Figure \ref{fig:gavgvsepoch} demonstrate the trend of $Gavg$ of two layers in a training with adaptive precision. $T_{min}$ is set to 1.0 in this demo. Layer A starts with a $Gavg$ below $T_{min}$, indicating it suffers quantisation underflow. APT allocates more bitwidth to lift the $Gavg$ of layer A above the threshold. Layer B is very easy to update at the beginning. Whenever $Gavg$ hits $T_{min}$, APT allocates more bitwidth to ensure layer B learns effectively. $T_{max}$ is set to $inf$ in this demo. It is also possible to use $T_{max}$ to reduce bitwidth for layers that do not suffer much from quantisation underflow.

\section{Experiments}

In this section we evaluate the APT with three datasets: CIFAR10 and CIFAR100 \cite{krizhevsky2009learning}. We follow the data augmentation in \cite{wu2018training} for training: 4 pixels are padded on each side, and a 32x32 patch is randomly cropped from the padded image or its horizontal flip. For testing, only single view of the original 32x32 image is evaluated. Three popular backbones, ResNet-20, ResNet-110 \cite{he2016deep}, and MobileNetV2 \cite{sandler2018mobilenetv2}, are included in the experiments.

Many works used sophisticated optimiser, such as Adam, in their experiments. We use SGD to show the potential of saving energy and memory usage. We use a weight decay of 0.0001 and momentum of 0.9, and adopt the weight initialization in \cite{he2015delving} and BN \cite{ioffe2017batch} with no dropout. The model is trained with a minibatch size of 128 on one GPU. For experiments on CIFAR10, we start with a learning rate of 0.1, divide it by 10 at 100 epoch and 150 epoch iterations, and terminate training at 200 iterations. For experiments on CIFAR100, we apply learning rate warm up, which reduces the learning rate down to 0.01 for the first two epochs and go back to learning rate scheduling for CIFAR10 for the rest of the training.

For all experiments we set initial bitwidth to 6, which will be explained later. Unless specified, we set the application specific parameter $(T_{min}, T_{max})$ to $(6.0, inf)$ for all experiments.  This setting is better for demonstrating the capability of APT in saving energy and memory usage. One can set $(T_{min}, T_{max})$ to other values that meets the requirements of the applications.

\subsection{Training Curves and Adaptive Precision Adjustment}

\begin{figure}
\centering
\includegraphics[width=0.9\linewidth]{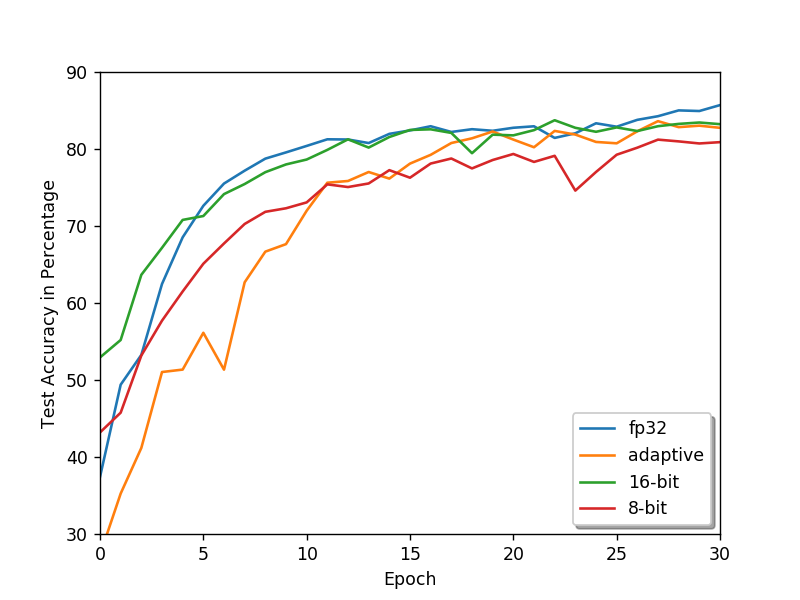}
\caption{Test Accuracy v.s. Epoch for ResNet20 on CIFAR10}
\label{fig:accvsepoch}
\end{figure}

We compare the our APT method with a vanilla SGD with different precision. The weights of all models are quantised for both forward pass and backward pass. Figure \ref{fig:accvsepoch} shows the training curve. 

fp32 and 16-bit model has the steepest learning curves among all, as they do not suffer much from quantisation underflow problem. The curve of 8-bit model does not climb as fast as fp32 and 16-bit does. An investigation into the training statistics shows that $Gavg$ of the all layers drop from the scale of 1 down to $1e-1$ within the first 50 epoch, which indicates quantisation underflow happens model wide and significantly slows down the training process of the 8-bit model.

On comparison, our adaptive training method starts with a model initialised with 6-bit weights. Its training curve starts with lower accuracy at the beginning, even lower than that of 8-bit. It overtakes the 8-bit and catch up with 16-bit and fp32. This is achieved by adaptively adjust layer-wise bitwidth of the model.

\begin{figure}
\centering
\includegraphics[width=0.9\linewidth]{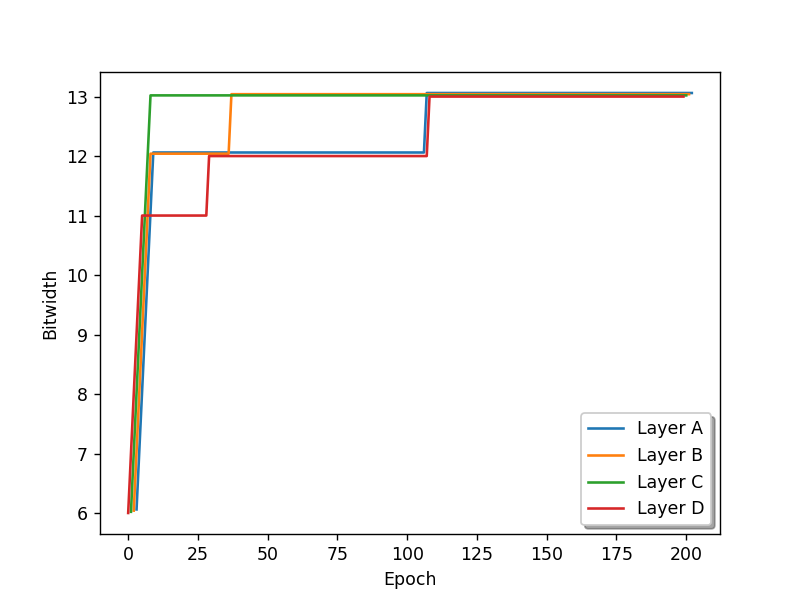}
\caption{Layer-wise Bitwitdh v.s. Epoch for ResNet20 on CIFAR10}
\label{fig:bitwidthvsepoch}
\end{figure}

Figure \ref{fig:bitwidthvsepoch} demonstrate the changes our method made to the layer-wise bitwidth of the model. Only four layers with weight parameters out of 20 are shown in this figure for the sake of clarity. Overlapping curved are shifted slight away from each other for the same purpose.

APT treats layers differently according to their learning effectiveness. Some layers are trained with lower bitwidth in early epochs and achieve solid energy savings. At 100 epoch, The precision of the first and last layer reach bitwidth of 13 as the training loss drops quickly after the decay of the learning rate.

Though Figure \ref{fig:bitwidthvsepoch} we understand that APT also serves as a neural architecture search method. It start training with an initial bitwidth of 6, which is for sure to be under-qualified with the target application. The adaptive training method will dynamically adjust the bitwidth of the network, and approaches the target accuracy with fewer energy. This means an initial bitwidth other than 6 leads to similar similar results. We choose 6 for all experiments just for demonstration purpose.

\subsection{Savings in Energy and Memory}

In this subsection we demonstrates that our adaptive training method saves training energy and memory. APT provides users trade-off between training energy and test accuracy. For this comparison. we train ResNet20 with CIRFAR10 dataset. The weights of all models are quantised for both forward pass and backward pass. The 12-bit, 14-bit, 16-bit and 32-bit models use fixed bitwidth throughout training process.  We do not include 10-bit, 8-bit or lower bitwidth because their suffer large accuracy loss and fall off charts in most cases. Our adaptive training method starts on a model with initial weight bitwidth of 6.

\begin{figure}
\centering
\includegraphics[width=0.9\linewidth]{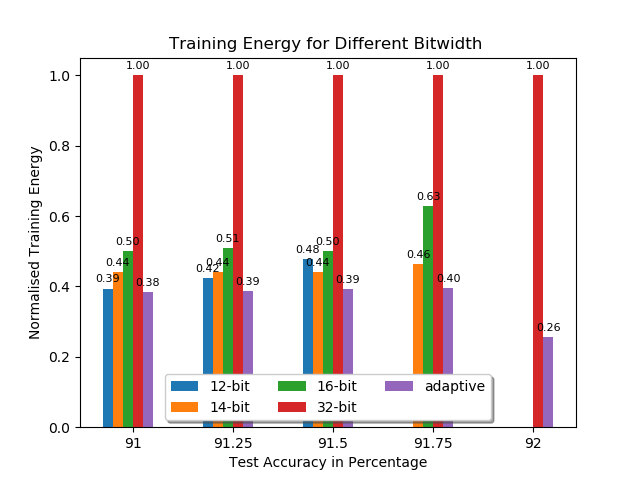}
\caption{Training Energy v.s. Bitwidth for ResNet20 on CIFAR10}
\label{fig:energyvsbitwidth}
\end{figure}

The grouped bars in Figure \ref{fig:energyvsbitwidth} show the energy our method saved given a target accuracy. X axis is the Top 1 accuracy ranging from 91\% to 92\%. All training energy cost are normalised to the cost of 32-bit model.

12-bit model has the least training energy cost among the all fixed-bitwidth models. It still spends 9\% more training energy than our adaptive method does. The 12-bit model is absent in the 91.75\% and 92\% group because within 200 epoch it is not able to reach an accuracy of 91.75\%.

In order to achieve the last few percentage of improvement in accuracy, models with fixed bitwidth spend a lot of training energy. For example, 16-bit model spends 13\% more energy to achieve an improvement of 0.25\% from 91.5\% to 91.75\%. On comparison, our adaptive method managed to achieve this by just 1\% additional energy.

One may notice that the training energy of our adaptive method drops from 40\% to 26\% when achieving the accuracy improvement from 91.75\% to 92\%. This is because the 32-bit model spends more training energy than our  method does for the same improvement. As the training energy in this figure is normalised to that of the 32-bit model, the training energy of our method drops from a percentage point of view.

In Figure \ref{fig:energyvsbitwidth}, APT trains with $T_{min}=6.0$. This hyper-parameter is application specific, which can be used as a trade-off between accuracy and training energy. Next we demonstrate that by tuning this threshold, one is able to achieve more savings in training energy and memory cost in the cost of accuracy loss.

\begin{figure}
\centering
\includegraphics[width=0.9\linewidth]{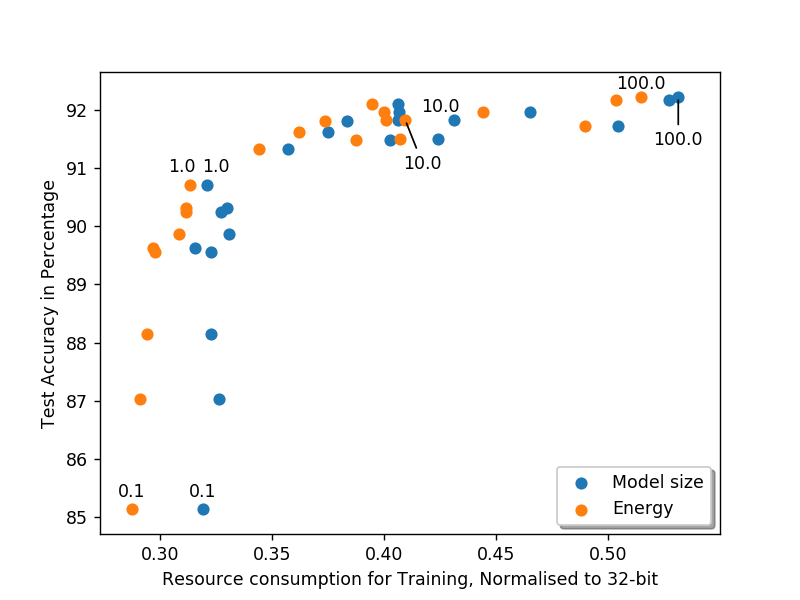}
\caption{Resource Consumption for Training v.s. Test Accuracy for ResNet20 on CIFAR10}
\label{fig:resourcevsaccuracy}
\end{figure}

In Figure \ref{fig:resourcevsaccuracy}, we use $Gavg$ threshold ranging from $0.1-100.0$ and generate a scatter plot of the training energy cost (in orange) for 200 epochs and corresponding accuracy the it achieves. All training energy cost are normalised to that of 32-bit model training. 

Through Figure \ref{fig:resourcevsaccuracy} we know that higher $Gavg$ threshold results in more training energy, memory usage and higher accuracy. For threshold below $1.0$, accuracy increases quickly along with the increase of training energy. The plateau appears on the right of threshold $1.0$ means extra training energy brings less improvements, which is very common during the training process. One can choose a $Gavg$ threshold that fits the need of the his or her applications.

Figure \ref{fig:resourcevsaccuracy} also presents the model size for training of the same experiment. We use low-precision representation of parameter during the back propagation in order to save memory usage for training. The memory usage is normalised to that of a 32-bit presentation. The memory usage follows the same trend as the training energy. 

\subsection{Comparison to Others}

Table \ref{tab:accuracy} shows the comparison to other quantisation methods. Model Precision in BPROP refers to the representation used in the update of weights. Many works \cite{hubara2017quantized, li2016ternary, zhu2016trained, zhou2016dorefa, wen2017terngrad, wang2019e2} keeps a fp32 copy of weights for back propagation in order to prevent a prolonged training process. There is no savings in memory usage for training. For example, TernGrad \cite{wen2017terngrad} is only for worker-to-server communication in distributed training, weights are still accumulated with fp32. In fact, \cite{wu2018training, yin2018binaryrelax} need more training epochs and/or sophisticated optimiser to reach comparable accuracy. With SGD optimiser and quantised model in BPROP, APT is able to achieve more than 50\% of savings in energy and memory usage with limited accuracy loss.

\begin{table*}[htb]
\caption{Comparison of network quantisation methods}
\begin{center}
\begin{threeparttable}
\begin{tabular}{ c c c c c c }
\hline
Method & Model Precision in BPROP & Optimizer & CIFAR10 & CIFAR100\\ 
\hline

BNN \cite{hubara2017quantized}  & FP32 & Adam & 89.85 & NA \\
TWN \cite{li2016ternary} & FP32 & BinaryRelax \cite{yin2018binaryrelax} & 88.65 (ResNet-20) \cite{yin2018binaryrelax} \tnote{b} & 68.95 (ResNet-110) \cite{yin2018binaryrelax} \tnote{b} \\
TTQ \cite{zhu2016trained}  & FP32 & Adam & 91.13 (ResNet-20) & NA \\
DoReFa Net \cite{zhou2016dorefa}  & FP32 & Adam & 89.52 \cite{zhou2019progressive} & 61.43 \cite{zhou2019progressive} \\

TernGrad \cite{wen2017terngrad}  & FP32 \tnote{a} & Adam & 85.64 (CifarNet) & NA \\

WAGE \cite{wu2018training}  & 8-bit & SGD & 93.22 (VGG-like) \tnote{b} & NA\\
E2 Train \cite{wang2019e2}  & FP32 & SGD & 93.07 (ResNet110) & NA \\
\hline
\textbf{APT}  & Adaptive & SGD & 92.23 (ResNet20) & 68.38 (ResNet110) \\
  &  &  & 93.96 (MobileNet v2) & \\
 \hline
\end{tabular}

\begin{tablenotes}
    \item[a] Only for worker-to-server communication in distributed training, weights are accumulated with float32
    \item[b] 300 training epochs
\end{tablenotes}
\end{threeparttable}
\end{center}
\label{tab:accuracy}
\end{table*}

\section{Conclusion and Future Work}

Training DNNs on resource constrained edge devices is challenging. We propose Adaptive Precision Training method that saves both energy cost and memory usage for training. APT evaluates how effective each layer learns with its current precision. Based on the evaluation, APT performs layer-wise precision adjustments dynamically to make sure the model learns effective throughout the training process. APT has a hyper-parameter $T_{min}$ which provides trade-off between training energy and accuracy. Tuning parameter $T_{min}$ requires application specific knowledge. In future, we are going to find automatic ways for choosing a proper $T_{min}$ in order to ease the use of APT.

\bibliographystyle{plain}
\bibliography{reference}

\end{document}